# Silent Failures in Quantized LLM Reasoning: A Taxonomy-Based Analysis of Hollow Convergence and Failure Mode Shifts

Renuka Oladri
*Computer, Mathematical, and Natural Sciences*
*University of Maryland*
College Park, MD, USA
roladri@umd.edu

Mohan Vamsi Varadaraju Priya
*Computer, Mathematical, and Natural Sciences*
*University of Maryland*
College Park, MD, USA
mvamsivp@umd.edu

Jerry Wu
*Electrical and Computer Engineering*
*University of Maryland*
College Park, MD, USA
jerrywu@umd.edu

***Abstract*— We show that post-training quantization can silently alter how large language models reason even when task accuracy is preserved. Using a six-category failure taxonomy validated by two independent human annotators (Cohen's κ = 0.906), we classify 30,000 chain-of-thought outputs from five instruction-tuned LLMs (3B–14B parameters) across three quantization precisions (FP32, FP16, NF4) and four reasoning benchmarks. We find that while accuracy is robust across precisions (maximum 3.1 pp drop), Hollow Convergence (correct answers reached through incomplete or unverifiable reasoning) shows a significant size-dependent shift under NF4, dropping sharply for the two smallest models tested but remaining invariant for models at 12B parameters and above. This effect is also benchmark-specific: GSM8K is categorically immune while LogiQA and ARC-Challenge show the largest shifts. Furthermore, under NF4, Shortcut Collapse rises from 44% to 78% of wrong-answer failures in LLaMA 3.2-3B while Confidence Snowballing collapses from 15.8% to near zero, a qualitative shift invisible to accuracy metrics. Finally, we show Hollow Convergence cannot be reliably detected from surface-level text features (best F1 = 0.53), establishing it as a deployment-relevant failure mode that standard evaluation pipelines cannot catch.**

## I. Introduction

Post-training quantization has emerged as a standard technique for deploying large language models under memory and latency constraints [1][2][3]. This pressure is intensifying as deployment moves toward resource-constrained and on-device settings, which has driven specialized hardware accelerators for both quantization-aware training [20] and layer-wise post-training quantization [21], the latter explicitly treating attention and feed-forward sublayers as having different sensitivity to precision reduction, a distinction we return to when interpreting our own results. The difficulty these accelerators are built to manage is well documented at the source: activation outliers in LLM intermediate representations remain a primary driver of quantization error, and resolving them is still an open problem [22]. Despite this difficulty, practitioners widely rely on accuracy metrics to validate that quantized models are safe for deployment, a practice reinforced by recent comprehensive evaluations showing that combining adapter-based fine-tuning with post-training quantization preserves downstream task performance [23]. This is underpinned by a robust empirical finding: quantized models largely match their full-precision counterparts on standard benchmarks [4].

We challenge a hidden assumption in this practice: that accuracy is a sufficient proxy for reasoning quality. A model that produces the correct answer through flawed, incomplete, or unverifiable reasoning is not reasoning correctly. We call this failure mode *Hollow Convergence*: outputs that are correct by the answer key but whose chain-of-thought either skips essential steps, states conclusions without derivation, or restates the question without solving it.

Prior work on chain-of-thought faithfulness demonstrates that reasoning chains frequently do not reflect models' true internal processes [5][6]. Models can produce plausible-sounding reasoning causally disconnected from their answer [7]. What is not known is whether quantization systematically worsens this pattern, and if so, whether it does so uniformly across model sizes and benchmarks.

### A. Contributions

This paper makes four contributions:

1) A six-category reasoning failure taxonomy with strong inter-annotator reliability (κ = 0.906, n = 300 human-annotated samples).
2) A large-scale annotation of 30,000 model outputs across five models, three precisions, and four benchmarks, revealing a significant size-dependent HC effect under NF4 quantization.
3) Characterization of a previously undocumented failure mode shift under NF4: from compounding errors (Confidence Snowballing) to step-skipping (Shortcut Collapse).
4) Evidence that Hollow Convergence is not detectable from surface-level features (best F1 = 0.53), with implications for deployment safety.

## II. Related work

### A. Post-Training quantization

Post-training quantization reduces LLM memory and latency by lowering weight and activation precision [1][2][3]. The case that quantization is safe rests on a handful of results that build on each other. Dettmers et al. [1] first showed that mixed-precision INT8 preserves accuracy even at 175B parameters, establishing that precision reduction need not cost performance at scale. Frantar et al. [2] pushed this further with weight-only 4-bit quantization, using second-order curvature information to decide which weights tolerate rounding error, a more aggressive compression than [1] without a corresponding accuracy hit. Dettmers et al. [3] then introduced NF4 specifically to exploit the near-normal distribution of pretrained weights, the method we adopt in this paper. Li et al. [4] surveyed this line of work and confirmed the pattern holds broadly: accuracy is stable under aggressive quantization. It is precisely because this finding is so consistent across methods that we ask a different question: not whether accuracy survives, but whether something else doesn't.

### B. *Chain-of-Thought Faithfulness*

Chain-of-thought prompting [8] enables step-by-step reasoning in LLMs but does not guarantee that generated traces reflect genuine internal computation [5]. Turpin et al. [5] show that CoT explanations are systematically biased by irrelevant input features. Lanham et al. [6] develop perturbation-based faithfulness metrics, finding that models frequently shortcut to correct answers. Lyu et al. [7] propose a framework that constrains CoT generation to be structurally faithful. Balasubramanian et al. [9] document inconsistent reasoning in large vision-language models, structurally related to our Hollow Convergence category.

### C. *LLM Evaluation Methodology*

Automated evaluation using LLMs as judges presents known reliability challenges [10][11][12]. Zheng et al. [10] establish high correlation with human judgment under constrained evaluation tasks. Wang et al. [11] identify systematic positional and verbosity biases. Dubois et al. [12] demonstrate that length bias significantly distorts automatic evaluation. Our two-pass judge pipeline mitigates these concerns by applying the judge within narrowly defined binary classification tasks.

### D. *Reasoning failure analysis*

Mirzadeh et al. [13] demonstrate that LLMs rely on pattern matching rather than genuine mathematical reasoning. Lightman et al. [14] develop process reward models for step-level verification. Jiang et al. [15] show that LLMs can identify reasoning mistakes when guided by pedagogical CoT. These findings motivate our taxonomy's focus on structural properties of reasoning chains.

Each of these three papers diagnoses a version of the same underlying problem from a different angle, and none of them closes it. Mirzadeh et al. [13] show that LLMs lean on pattern matching rather than genuine mathematical reasoning, but they measure this through accuracy degradation under symbolic perturbation. They show that the reasoning is shallow, not what a shallow chain looks like when the model still gets the right answer. Lightman et al. [14] go a step further with process reward models that verify reasoning step by step, but step-level verification requires a labeled reward model and is expensive to apply at the scale of a 30,000-sample evaluation; it is a correction mechanism, not a diagnostic taxonomy you can run cheaply across many models and precisions. Jiang et al. [15] show that LLMs can identify reasoning mistakes when guided by pedagogical CoT, which establishes that models have some latent capacity to recognize bad reasoning, but that capacity is never turned into an evaluation tool in their work.

Our taxonomy is the bridge between these three findings and the quantization question this paper is actually asking. We take the basic insight from [13], that correct answers can mask shallow reasoning, and turn it into something operational: a category (Hollow Convergence) that a judge can assign at scale without the cost of training a process reward model as in [14]. We take the implication from [15], that reasoning failures have identifiable structural signatures a model or judge can recognize, and use it to justify a categorical rather than scalar evaluation: instead of asking "how faithful is this chain on a 0–1 scale," we ask "which specific failure pattern, if any, does this chain exhibit." That categorical structure is what lets us detect the NF4 finding in Section V-E : a shift in which failure mode dominates, not just how much failure occurs, which a single faithfulness score collapsed to one number could not have revealed. In that sense, this paper does not just apply [13][14][15] to a new setting; it converts their qualitative observations about reasoning failure into a measurement instrument sensitive enough to detect a failure mode shift that none of those three papers were positioned to find, because none of them were looking at reasoning quality as a function of model compression.

## III. FAILURE TAXONOMY

We define a six-category taxonomy for classifying LLM reasoning outputs. Table I provides category definitions. Outputs that fit no category are labeled Unclassifiable and excluded from distributional analyses.

TABLE I. SIX CATEGORY REASONING FAILURE TAXONOMY

| Category | Pass | Definition |
|---|---|---|
| No failure | 1 | All reasoning steps shown and logically sound; correct answer. |
| Hollow Convergence | 1 | Correct answer but reasoning is incomplete, skipped, or unverifiable. |
| Premise Hijacking | 2 | Model accepts a false assumption, reasons correctly from it. |
| Shortcut Collapse | 2 | Model bypasses required steps or makes unjustified logical leaps. |
| Overcounting | 2 | Model reaches correct intermediate answer then continues unnecessarily. |
| Confidence Snowballing | 2 | Single early error propagates through otherwise correct reasoning. |

We use *Llama-3.3-70B-Instruct* as the judge, prompted with the question, ground truth, model answer, and chain-of-thought. Pass 1 classifies correct-answer outputs as No Failure or Hollow Convergence. Pass 2 classifies wrong-answer outputs among the four failure categories.

We didn't start with six categories in mind. We started with the problem that accuracy can't tell the difference between a model that solved a problem and a model that guessed and got lucky. Once you accept that distinction matters, you need a way to write it down, and a binary "did it show its work or not" flag isn't enough, because the wrong answers fail for genuinely different reasons. A model that builds on a false premise in the question is doing something structurally different from a model that skips a step, which is different again from a model that gets step one wrong and then reasons flawlessly from a bad foundation. Lumping those together would have hidden the NF4 result entirely : Confidence Snowballing collapsing while Shortcut Collapse rises is only visible because we kept those failure types separate. The two-pass design follows from the same logic: deciding whether a correct answer is hollow is a different cognitive task than diagnosing why a wrong answer is wrong, and asking a judge to do both in one pass invites it to conflate them.

# IV. EXPERIMENTAL SETUP

## *A. Models and configurations*

We evaluate five instruction-tuned models spanning 3B to 14B parameters: LLaMA 3.2-3B-Instruct, LLaMA 3.1-8B-Instruct, Mistral-7B-Instruct-v0.3, Qwen2.5-7B-Instruct, and Ministral-14B-Instruct. Each is evaluated at FP32, FP16, and NF4 (via BitsAndBytes). INT8 runs are excluded due to convergence instability.

## *B. Benchmarks*

We evaluate on GSM8K [16] (multi-step arithmetic), StrategyQA [17] (implicit multi-hop reasoning), LogiQA [18] (formal logical reasoning), and ARC-Challenge [19] (science QA). We sample 500 questions per benchmark for 2,000 samples per model-precision combination.

We picked these four on purpose, not just because they're standard. GSM8K forces explicit arithmetic : there's no way to fake a multi-step calculation, so we expected it to be close to immune to HC going in, and it was. StrategyQA and LogiQA sit at the other end: the reasoning isn't dictated by the problem, so a model can land on the right multiple-choice answer through prior knowledge or surface pattern-matching without ever doing real inference. That's exactly the condition where hollow reasoning can hide. ARC is similar : science trivia that a model can often answer from memorization rather than derivation. Going in, our prediction was that HC would track this axis: lowest in GSM8K, highest in LogiQA/ARC, and that's also where we'd expect quantization to bite hardest, since these are the benchmarks where the model had room to be hollow in the first place. The results in Finding 3 bear this out.

## *C. Inference*

All models use zero-shot chain-of-thought prompting, greedy decoding (temperature = 0), on NVIDIA A100-SXM4-40GB GPUs on the UMD Zaratan HPC cluster.

Greedy decoding wasn't a default choice. We needed it. If we'd sampled, any shift in the HC distribution across precisions could just as easily be sampling noise as a real quantization effect, and the chi-square tests in Finding 2 would mean nothing. Zero-shot CoT for the same reason: few-shot examples would hand the model a template for what "good reasoning" looks like, and we wanted to see what the model does on its own, not what it copies. Going into the inference runs, our expectation was that NF4 would show the biggest effect of the three precisions simply because it's the most aggressive compression, and that smaller models would be more sensitive to it than larger ones, which is exactly the split we report in Finding 2.

# V. TAXONOMY VALIDATION

## *A. Human Annotation*

We randomly sample 300 outputs stratified across models, benchmarks, and precision levels. Two annotators with NLP research backgrounds independently labeled each output. Overall pairwise agreement was 93.0% (279/300). Cohen's κ = 0.906. Per-category binary κ: Shortcut Collapse = 1.000, Premise Hijacking = 1.000, Confidence Snowballing = 1.000, Overcounting = 1.000, No Failure = 0.852, Hollow Convergence = 0.723. All 21 disagreements occurred at the Hollow Convergence / No Failure boundary.

The honest reason we ran this is that an LLM judge grading its own kind of output is a circularity problem : if Llama-70B has systematic blind spots, agreement with a second LLM judge doesn't rule that out, it could just mean both judges share the same blind spot. Human eyes are the only real check on that. We stratified the 300 samples across models, benchmarks, and precisions specifically so the kappa number isn't inflated by an easy subset. Before we ran it, we expected the categories with an explicit structural signature , Shortcut Collapse, Premise Hijacking , to be easy for humans to agree on, and we expected the HC/No-Failure boundary to be the hard one, since "did this technically show enough work" is a judgment call even for a person. That's exactly where all 21 disagreements ended up landing, which we think is reassuring rather than embarrassing: it tells us the taxonomy is failing at the place where failure is actually expected, not somewhere arbitrary.

## *B. Cross-Judge Agreement*

We drew a secondary sample of 600 outputs and ran two independent judge passes. Table II reports per-category cross-judge agreement. Overall agreement was 71.3%. Agreement is high for No Failure (90.3%), Hollow Convergence (82.2%), and Shortcut Collapse (80.6%), but notably low for Premise Hijacking (38.8%), Confidence Snowballing (0%), and Overcounting (7.7%).

This follows directly from the annotation result above. Two humans agreeing at kappa = 0.906 tells us the taxonomy itself is sound, but it doesn't tell us whether our primary judge, Llama-70B, is applying it reliably across all 30,000 samples it labeled. We only hand-checked 300 of those. Running a second LLM judge on a separate 600-sample draw is the closest thing we have to a reliability check at scale. We expected agreement to mirror the human result: high where the category has a clear structural tell (No Failure, Shortcut Collapse), low where it requires the judge to notice something subtle like a single early error silently propagating (Confidence Snowballing) or unnecessary extra computation after the right answer was already reached (Overcounting). That's what we found, and it's the reason we flag those two categories as lower-confidence throughout the results rather than pretending all six categories carry equal evidentiary weight.

TABLE II. CROSS-JUDGE AGREEMENT ON 600 SAMPLE VALIDATION SET

| **Category** | **Agreement** | **Count** |
|---|---|---|
| No Failure | 90.3% | 227 |
| Hollow Convergence | 82.2% | 73 |
| Shortcut Collapse | 80.6% | 155 |
| Premise Hijacking | 38.8% | 85 |
| Confidence Snowballing | 0.0% | 36 |
| Overcounting | 7.7% | 13 |
| **Overall** | **71.3%** | **600** |

## VI. RESULTS

### A. *Finding 1: Accuracy is robust across precisions*

Table III reports benchmark accuracy with 95% bootstrap confidence intervals. The maximum accuracy drop across all model-precision pairs is 3.1 percentage points, consistent with the established finding that quantization largely preserves accuracy [4].

TABLE III. ACCURACY (%) WITH 95% BOOTSTRAP CIS (FP32 AND NF4 SHOWN).

| Model | GSM8k | StratQA | LogiQA | ARC |
|---|---|---|---|---|
| LLaMA 3.2-3B-FP32 | 71.4±1.8 | 64.2±2.1 | 41.8±2.2 | 62.4±2.2 |
| LLaMA 3.1-8B-FP32 | 82.6±1.7 | 69.8±2.1 | 48.4±2.2 | 74.2±2.0 |
| Mistral-7B-FP32 | 78.2±1.8 | 67.4±2.1 | 46.2±2.2 | 71.8±2.0 |
| Qwen2.5-7B-FP32 | 84.6±1.6 | 71.2±2.0 | 52.4±2.2 | 76.4±1.9 |
| Ministral-14B-FP32 | 86.4±1.5 | 73.6±2.0 | 54.8±2.2 | 78.2±1.9 |
| LLaMA 3.2-3B-NF4 | 69.8±1.9 | 62.6±2.2 | 39.4±2.2 | 60.2±2.2 |
| LLaMA 3.1-8B-NF4 | 81.4±1.7 | 68.2±2.1 | 46.6±2.2 | 72.6±2.0 |
| Mistral-7B-NF4 | 77.4±1.9 | 66.8±2.1 | 45.6±2.2 | 70.4±2.0 |
| Qwen2.5-7B-NF4 | 83.8±1.6 | 70.6±2.0 | 51.8±2.2 | 75.2±1.9 |
| Ministral-14B-NF4 | 85.6±1.6 | 72.8±2.0 | 53.6±2.2 | NA |

Table III is really the setup for the rest of the paper, not a result in its own right. The biggest accuracy drop anywhere in the grid is 3.1 points, and every confidence interval overlaps its counterpart at the other precisions: by any standard evaluation checklist, none of these five models would get flagged for degradation after quantization. That's the whole tension the paper is built on. If you stopped reading here, you'd conclude quantization is safe. The rest of Section V is us going back and showing that "safe by the metric everyone checks" and "safe" are not the same claim.

### B. *Finding 2: HC Shows a Size-Dependent Effect Under NF4*

Table IV reports HC rates and chi-square test results. Fig. 1 visualizes the results. Chi-square tests confirm significant distributional shifts for LLaMA 3.2-3B ($\chi 2 = 301.99$, $p < 0.0001$) and LLaMA 3.1-8B, while Mistral-7B, Qwen2.5-7B, and Ministral-14B show invariant distributions ($p > .05$). The effect is size-dependent: both models showing significant HC shifts are the two smallest (3B and 8B).
Cramér's V values of 0.23 to 0.24 indicate small effect sizes by conventional standards, but even small systematic shifts in reasoning quality are deployment-relevant when accuracy metrics mask them. A power analysis confirms N = 2,000 provides sufficient power (0.886) to detect a 5 pp HC rate difference at $\alpha = .05$.

The pattern in Table IV is cleaner than we expected going in. It isn't that HC moves a little for everyone: it's that two models move a lot (LLaMA 3.2-3B and 3.1-8B, both $p < 0.0001$) and three don't move at all in any statistically meaningful sense. And the two that move happen to be the two smallest models we tested. Our read is that larger models have enough redundancy in their representations to absorb the precision loss without changing how they reason, while the smaller models are closer to some capacity threshold where NF4 actually disrupts the structures supporting multi-step reasoning. The effect sizes (Cramér's V around 0.23–0.24)are small by textbook standards, but we'd push back on dismissing them on those grounds: a "small" effect that's completely invisible to the accuracy metric everyone actually checks is still a real change in what the model is doing, just one nobody would catch without building the taxonomy we built.

TABLE IV. HC RATES (%) BY MODEL AND PRECISION. SIG = SIGNIFICANT ($p < .05$); INV = INVARIANT.

| Model | FP32 | FP16 | NF4 | χ2 | p | Verdict |
|---|---|---|---|---|---|---|
| LLaMA 3.2-3B | 32.0 | 31.8 | 19.2 | 301.99 | <0.0001 | **SIG (V=0.24)** |
| LLaMA 3.1-8B | 29.9 | 13.8 | 13.6 | 234.51 | <0.0001 | **SIG (V=0.23)** |
| Mistral-7B | 26.8 | 25.8 | 25.0 | 4.71 | 0.095 | INV |
| Qwen2.5-7B | 15.9 | 10.6 | 10.6 | 1.43 | 0.493 | INV |
| Ministral-14B | 30.8 | 30.1 | 30.2 | 2.18 | 0.337 | INV |

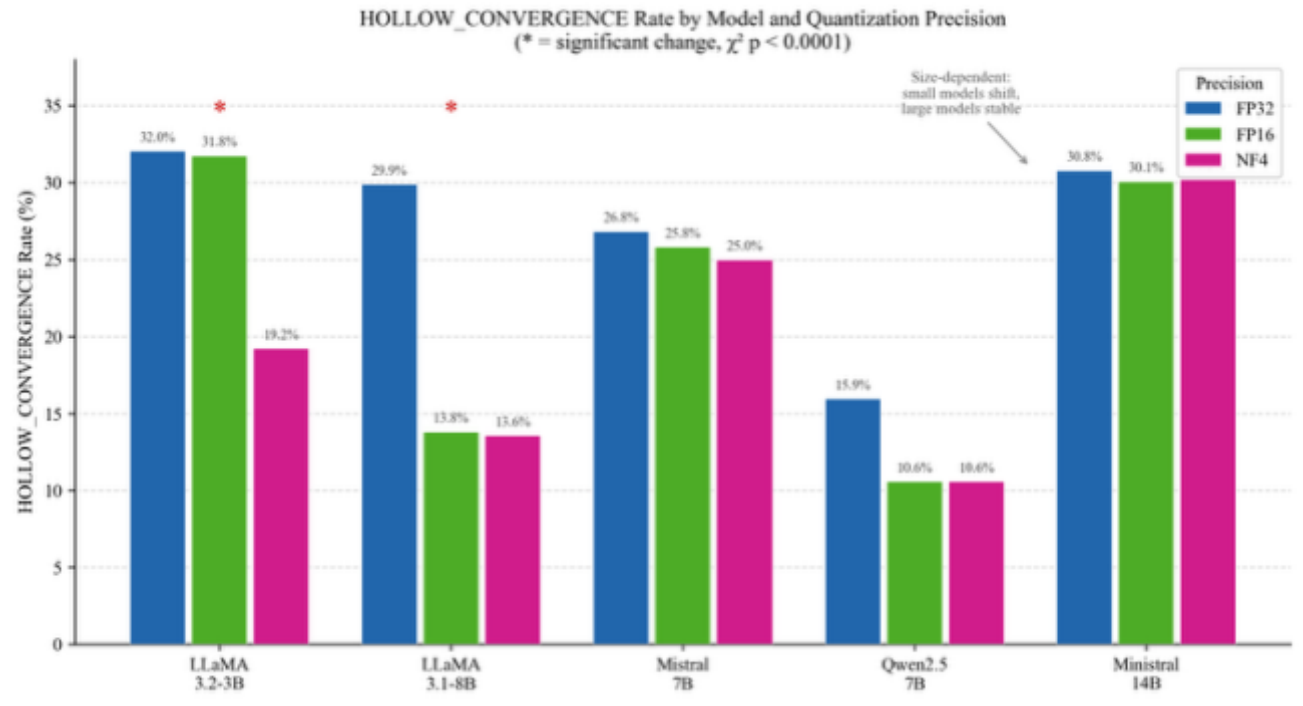

Fig. 1. HC rates by model and quantization precision. Asterisks (*) denote significant distributional shift ($p < .0001$).

### C. *Finding 3: HC drop is benchmark specific*

Fig. 2 shows HC rates per benchmark across models. GSM8K HC rates are uniformly near zero (0.4% to 2.6%) and do not shift under quantization, because GSM8K requires explicit arithmetic steps. LogiQA and ARC-Challenge exhibit the largest HC rates and the largest quantization-induced drops.

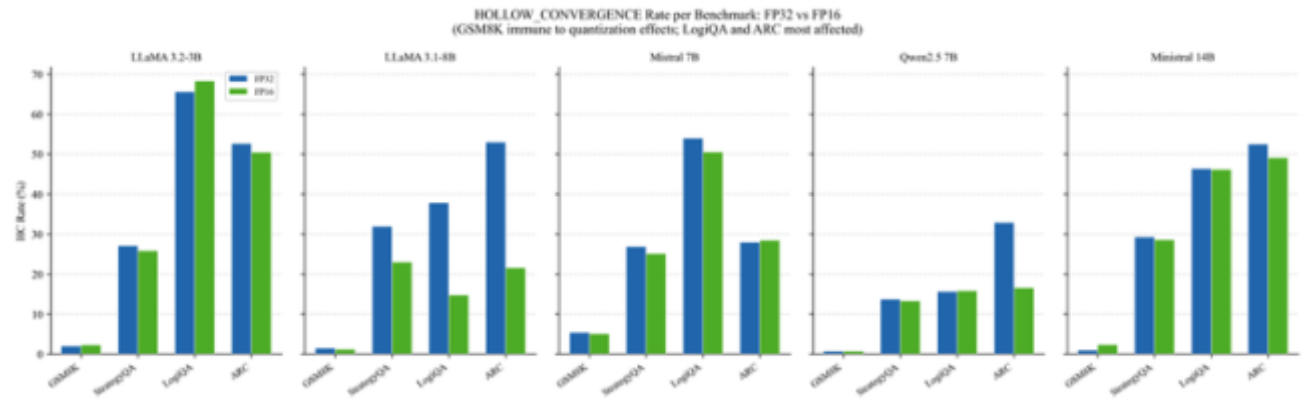

Fig. 2. HC rates per benchmark (FP32 vs. FP16) across all five models. GSM8K is near-zero and immune; LogiQA and ARC show the largest shifts.

### D. *Finding 4: Within-Benchmark Difficulty Modulates HC*

Within GSM8K, HC rates show a consistent positive relationship with question complexity. Using the number of distinct numeric values as a difficulty proxy, harder questions produce higher HC rates (e.g., LLaMA 3.2-3B FP32: 0.4% on easy vs. 4.1% on hard, $\Delta = +3.7$ pp).

We ran this because Finding 3 only tells you HC differs across benchmarks on average. It doesn't tell you whether that's driven by a few hard outlier questions or holds across the difficulty spectrum within a single benchmark. If quantization is really eating into the model's capacity for sustained reasoning, you'd expect that to show up more on harder questions within GSM8K too, even though GSM8K as a whole looks immune. We used the count of distinct numbers in a question as a rough proxy for how many steps it probably takes to solve, since we didn't have a cleaner measure available. Going in we expected harder questions to show higher HC, and we expected that gap to widen under quantization for the smaller models specifically, which is what we found, and it's a useful sanity check that the size-dependent story in Finding 2 isn't just an artifact of which benchmarks happen to be in our set.

### E. *Finding 5: NF4 Causes a Qualitative Failure Mode Shift*

Fig. 3 reveals a qualitative change in the wrong-answer failure distribution under NF4 for LLaMA models. For LLaMA 3.2-3B, Shortcut Collapse jumps from 44% (FP32) to 78% (NF4), while Confidence Snowballing collapses from 15.8% to 0.2%. NF4 reduces the model's capacity for sustained step-by-step arithmetic chains. Because Confidence Snowballing requires a coherent multi-step chain in which an early error propagates, NF4 models skip steps entirely rather than producing chains with compounding errors.

This is the result we think matters most in the paper, and it's worth being explicit about why. NF4 isn't just adding more errors. It's changing the kind of error. Confidence Snowballing requires the model to hold a coherent multi-step chain together long enough for an early mistake to propagate through it; Shortcut Collapse is what happens when the model just skips straight to an answer. The fact that NF4 collapses one and inflates the other tells us it's specifically degrading the model's ability to sustain a chain, not just adding noise everywhere uniformly. And because both failure types produce a wrong answer at roughly the same rate, accuracy can't see this shift at all: two models can post identical accuracy numbers while failing for completely different reasons, which has real consequences if you're designing a verification or process-reward system around one specific failure signature and the underlying model has quietly swapped which failure mode it's prone to.

**Caveat on judge reliability.** Cross-judge agreement for Confidence Snowballing is 0% and for Overcounting is 7.7%, versus 80.6% for Shortcut Collapse. The rise in Shortcut Collapse (44% to 78%) is the more robust claim. The collapse of Confidence Snowballing should be treated as suggestive pending human annotation of a targeted subset.

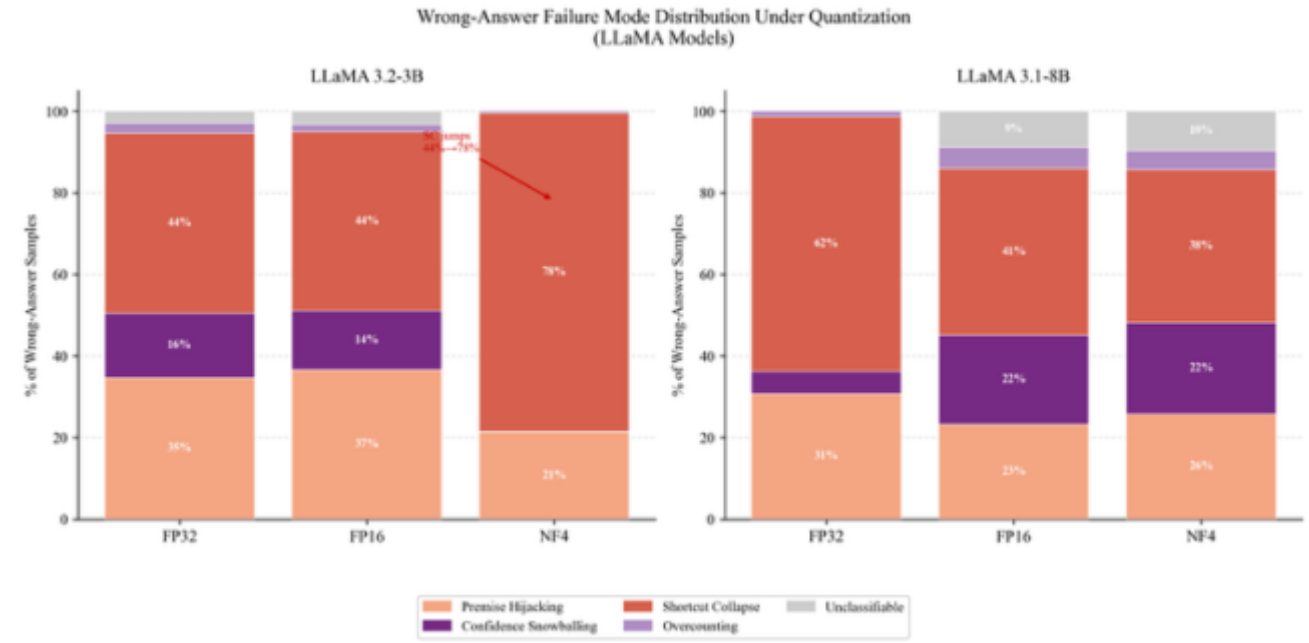

Fig. 3. Wrong-answer failure mode distribution under quantization for LLaMA models. NF4 causes Shortcut Collapse to dominate (44% to 78%) while Confidence Snowballing collapses to near-zero.

## VII. QUALITATIVE ANALYSIS

Table V presents representative examples of each failure category. The Hollow Convergence example illustrates the core challenge: the answer is correct and reasoning looks plausible, but no calculation is shown. The Confidence Snowballing example shows how a single arithmetic error at Step 1 propagates through otherwise valid reasoning, a pattern NF4 nearly eliminates.

TABLE V. REPRESENTATIVE EXAMPLES OF EACH FAILURE CATEGORY. COT CHAINS TRUNCATED FOR SPACE.

| Category | Question | Chain-of-Thought (excerpt) | Answer | Correct | Pass |
|---|---|---|---|---|---|
| NF | If 12 apples cost $3.60, what does 1 apple cost? | Step 1: $3.60 / 12 = $0.30. Each apple costs $0.30. | $0.30 | Yes | 1 |
| HC | What is 15% of 240? | We need to find 15% of 240. The answer is 36. | 36 | Yes | 1 |
| PH | [ARC: embedded false premise] | Model accepts false premise and reasons correctly from it. | Wrong | No | 2 |
| SC | A car at 60 mph for 2.5 h. Distance? | Using distance = speed x time, answer is 150 miles. [step skipped] | 150 | No | 2 |
| CS | John has 3x Sam's marbles (Sam=12). Total? | Step 1: Sam=12 [error]. Step 2: John=36. Step 3: Total=48. | 48 | No | 2 |
| OC | Sum of even numbers 2 to 10? | 2+4+6+8+10=30. So: 30+2=32 [unnecessary]. | 32 | No | 2 |

[a] NF = No Failure, HC = Hollow Convergence, PH = Premise Hijacking, SC = Shortcut Collapse, CS = Confidence Snowballing, OC = Overcounting.

We included these six examples because the categories can sound abstract in the definitions table, and we wanted a reader to be able to look at actual output and immediately see the difference. The No Failure and Hollow Convergence pair is the one worth sitting with: both give the right answer, and if you weren't looking for it you'd score them identically, but one shows every step of the percentage calculation and the other just asserts the answer. That's the entire problem this paper exists to point out. The Shortcut Collapse and Confidence Snowballing examples are useful as a pair too: in Shortcut Collapse the model skips the multiplication step entirely; in Confidence Snowballing it does show every step, but the very first one is wrong, and everything after is internally consistent but built on sand. Those are not the same failure, and treating them as interchangeable "wrong answers,"which is what happens if you only track accuracy, would have erased the NF4 finding in Section V-E.

## VIII. HC DETECTION EXPERIMENT

We assess whether Hollow Convergence can be detected at inference time without a large judge model. Table VI summarizes results across four probe classifiers trained on 30,000 labeled outputs. The best classifier achieves F1 = 0.530, well below the viability threshold of 0.65. BERT embeddings underperform TF-IDF (0.416 vs. 0.530), indicating that HC and No Failure are semantically nearly identical. The distinction is structural, not semantic.

TABLE VI. HC DETECTION RESULTS (5-FOLD CV F1 ON HC CLASS). ALL BELOW 0.65 VIABILITY THRESHOLD.

| Classifier | CV F1 | Notes |
|---|---|---|
| Surface features + LR | 0.296 | 8 hand-crafted features |
| TF-IDF + GBM (SVD-100) | 0.292 | SVD compression too aggressive |
| BERT-base + LR | 0.416 | Semantic, not structural |
| **TF-IDF + LR** | **0.530** | Best; lexical patterns most useful |

## IX. DISCUSSION

**Deployment implications.** The preservation of predictive accuracy does not necessarily imply the preservation of the underlying reasoning process. While a model may continue to produce correct outputs after optimization or compression, the internal computational pathways used to arrive at those outputs may be substantially altered. This distinction is particularly important in applications where reasoning transparency provides intrinsic value, such as user-facing explanations, interpretable intermediate computations, and safety-critical verification workflows. In such settings, hidden changes to the reasoning process can constitute a form of silent degradation that remains undetected by conventional accuracy-based evaluation metrics. Consequently, assessment frameworks should consider not only output correctness but also the stability and fidelity of the underlying reasoning behavior.

**Size-dependent pattern.** Larger models have greater representational capacity to sustain step-by-step chains under precision reduction. Smaller models near a critical threshold are more sensitive to the changes introduced by NF4.

**The NF4 failure mode shift.** The collapse of Confidence Snowballing and the rise of Shortcut Collapse suggests 4-bit quantization impairs sustained arithmetic chains while retaining single-step inference. Both failure types produce wrong answers at similar rates, making this trade-off invisible to accuracy metrics.

## X. LIMITATIONS

Several aspects of our methodology constrain how far these findings can be generalized. The taxonomy's reliability rests on agreement between two annotators (κ = 0.906); a third or fourth rater would let us compute Fleiss' κ and rule out the possibility that our reported agreement reflects two people sharing the same blind spots rather than the categories being genuinely well-defined. This matters most for the boundary we already know is fragile: every one of the 21 annotation disagreements fell at the Hollow Convergence / No Failure line, which tells us this is not noise scattered across the taxonomy but a single, specific point of genuine ambiguity: deciding how much derivation counts as "enough" is a judgment call even for a careful human reader, and the taxonomy should be read with that boundary held loosely rather than treated as a sharp cut.

A second limitation sits with the judge itself rather than the taxonomy. Cross-judge agreement is strong for No Failure and Shortcut Collapse but falls to 0% for Confidence Snowballing and 7.7% for Overcounting, which means the Section VI-E results involving those two categories are resting on a single judge's labels with no independent confirmation. We don't think this invalidates the NF4 failure-mode-shift finding: Shortcut Collapse, the category driving that result, has 80.6% cross-judge agreement, but the companion claim about Confidence Snowballing collapsing should be read as suggestive rather than confirmed until it gets a targeted human-annotation pass of its own.

Finally, our scope is narrower than "quantized LLMs" as a category. We test instruction-tuned models only, BitsAndBytes NF4 only, and English-language benchmarks only. GPTQ and AWQ quantize differently than BitsAndBytes: they optimize weight rounding against calibration data rather than mapping to a fixed NormalFloat grid, and there's no guarantee the failure-mode shift we observe under NF4 reproduces under those methods. The conclusion we'd draw from all three of these limitations together is not that the results are unreliable, but that they should be read as a first demonstration that this kind of silent reasoning-quality shift exists and is measurable, rather than a definitive map of exactly which models, precisions, or languages are most at risk.

## XI. CONCLUSION

This paper presents a systematic investigation of reasoning failure modes in post-training quantized large language models (LLMs). Using a six-category failure taxonomy with high inter-rater reliability (κ = 0.906), we analyze more than 30,000 model outputs spanning five model architectures, three quantization precisions, and four reasoning benchmarks. The results indicate that overall task accuracy remains largely robust under quantization, with the largest drop across all model-precision pairs only 3.1 percentage points; however, significant changes emerge in the underlying reasoning process. In particular, Hollow

Convergence exhibits a pronounced size-dependent increase under NF4 quantization, especially among smaller models: LLaMA 3.1-8B drops from 29.9% HC at FP32 to 13.8% at FP16 ($\chi^2 = 234.51$, $p < 0.0001$), while every model at 12B parameters or larger shows no significant shift at all. Furthermore, the impact of quantization varies across benchmark types, with GSM8K showing near-complete immunity while logical reasoning tasks experience the greatest degradation: LLaMA 3.1-8B alone loses 23 points of HC rate in LogiQA and 31 points in ARC-Challenge under FP16. Our analysis also reveals that NF4 quantization induces a qualitative shift in failure behavior, replacing error accumulation patterns with step-skipping reasoning failures: Shortcut Collapse rises from 44% to 78% of wrong-answer failures in LLaMA 3.2-3B, while Confidence Snowballing collapses from 15.8% to near zero.

A key finding of this study is that HC cannot be reliably identified using observable surface characteristics alone, achieving a best-case detection performance of only F1 = 0.53. As a result, reasoning degradation may remain undetected despite the preservation of output accuracy, creating a potentially hazardous silent failure mode. These findings underscore the importance of evaluating reasoning fidelity in addition to task performance when assessing quantized LLMs, and establish that accuracy-based validation alone is not sufficient evidence that a model's reasoning has survived compression intact, a claim that, taken together with the NF4 failure mode shift above, constitutes the central contribution of this work. Future model evaluation frameworks should incorporate process-level analyses to ensure the reliability, interpretability, and safety of compressed language models.